\documentclass[11pt,a4paper]{article}
\usepackage[hyperref]{acl2021}
\usepackage{times}
\usepackage{latexsym}
\usepackage{makecell}

\usepackage{microtype}

\aclfinalcopy 
\def\aclpaperid{477} 

\usepackage{booktabs}
\usepackage{epsfig}
\usepackage{graphicx}
\usepackage{xspace}
\usepackage{amsmath}
\usepackage{colortbl}
\usepackage{subcaption}
\usepackage{multirow}
\usepackage[T1]{fontenc}

\usepackage[utf8]{inputenc}

\usepackage{microtype}

\newcommand{\sys}{IrEne\xspace} 

\definecolor{qcolor}{HTML}{cc0066}

\title{\sys: Interpretable Energy Prediction for Transformers}

\author{Qingqing Cao, Yash Kumar Lal, Harsh Trivedi, \\{\bf Aruna Balasubramanian, Niranjan Balasubramanian} \\
Department of Computer Science\\
Stony Brook University\\
Stony Brook, NY 11794, USA \\
\texttt{\{qicao,ylal,hjtrivedi,arunab,niranjan\}@cs.stonybrook.edu}
}

\begin{document}
\maketitle

\setlength{\abovedisplayskip}{2pt}
\setlength{\belowdisplayskip}{2pt}

\begin{abstract}

Existing software-based energy measurements of NLP models are not accurate because they do not consider the complex interactions between energy consumption and model execution. We present \sys, an interpretable and extensible energy prediction system that accurately predicts the inference energy consumption of a wide range of Transformer-based NLP models. 
\sys\ constructs a model tree graph that breaks down the NLP model into modules that are further broken down into  low-level machine learning (ML) primitives. \sys\ predicts the inference energy consumption of the ML primitives as a function of generalizable features and fine-grained runtime resource usage. \sys\ then aggregates these low-level predictions recursively to predict the energy of each module and finally of the entire model. Experiments across multiple Transformer models show \sys\  predicts inference energy consumption of transformer models with an error of under 7\% compared to the ground truth. In contrast, existing energy models see an error of over 50\%. We also show how  \sys\  can be used to conduct energy bottleneck analysis and to easily evaluate  the energy impact of different architectural choices. We release the code and data at \url{https://github.com/StonyBrookNLP/irene}.
\end{abstract}

\section{Introduction}
Accurately measuring the energy consumption of NLP models is becoming ever more important. Models are growing exponentially, with billions, even approaching trillions, of parameters with correspondingly large resource consumption (e.g. GPT-3~\cite{brown2020LanguageModels} has 175 billion parameters and Switch Transformers can have 1.6 trillion parameters~\cite{fedus2021SwitchTransformers}
). Recent works have sought to estimate energy consumption and suggest ways to reduce the resulting costs and carbon impacts~\cite{strubell2019EnergyPolicy,schwartz2019GreenAIa,henderson2020SystematicReporting,anthony2020CarbontrackerTracking}

Unfortunately, there are no easy-to-use and accurate solutions for measuring or predicting the energy consumption. On the one hand, measuring energy consumption directly through hardware power monitors is not feasible as it requires exclusive access to the hardware and detailed instrumentation. On the other hand, there are software models that predict energy as a function of resource utilization~\cite{strubell2019EnergyPolicy,henderson2020SystematicReporting} but these energy prediction models are inaccurate~\cite{cao-etal-2020-towards}. The inaccuracy stems from the prediction models not accounting for the complex interactions between energy consumption and resource utilization. 

In this work, we focus on inference energy which can incur substantial costs especially for models that support high-volume web services. We ask how we can build an energy prediction method that is accurate, interpretable, and extensible. We make three  contributions in answering this question.

First, we frame the problem of interpretable energy prediction over a {\em model tree} abstraction. This abstraction represents the model as the root node that is composed from model-specific modules, which themselves are recursively composed from lower-level machine learning (ML)  primitives, ones that are not model-specific. Given a model, the energy prediction problem is framed as the task of predicting the energy of all the nodes in its model tree abstraction. The result is that \sys\ can predict not only the inference energy consumption of the entire model, but also of its components, making the energy prediction highly interpretable.

Second, we develop \sys, that includes a multi-level prediction method that predicts energy in all nodes of the abstraction tree in a bottom-up fashion using resource utilization and model description features. For each of the leaf-nodes that are re-used in different models, the ML primitives, \sys\ uses a separate regressor trained on ground-truth energy measurements. One simple way to get energy for all other higher-level nodes is to recursively sum-up the values. While this works reasonably well (even better than a prior prediction model), direct summing of the raw predictions is sub-optimal because the error can propagate through the model tree thus making upper-level nodes estimation more erroneous.
Instead, we learn a single regressor for all intermediate nodes, one that essentially adjusts the sum of children's predicted energy values based on features of the children. Since \sys\  is built on top of energy predictions of ML primitives that are not model specific, it is generalizable and can be used to predict the energy for previously unseen (Transformer-based) models.



\begin{figure*}[ht!]
  \centering
		\includegraphics[width=0.7\linewidth]{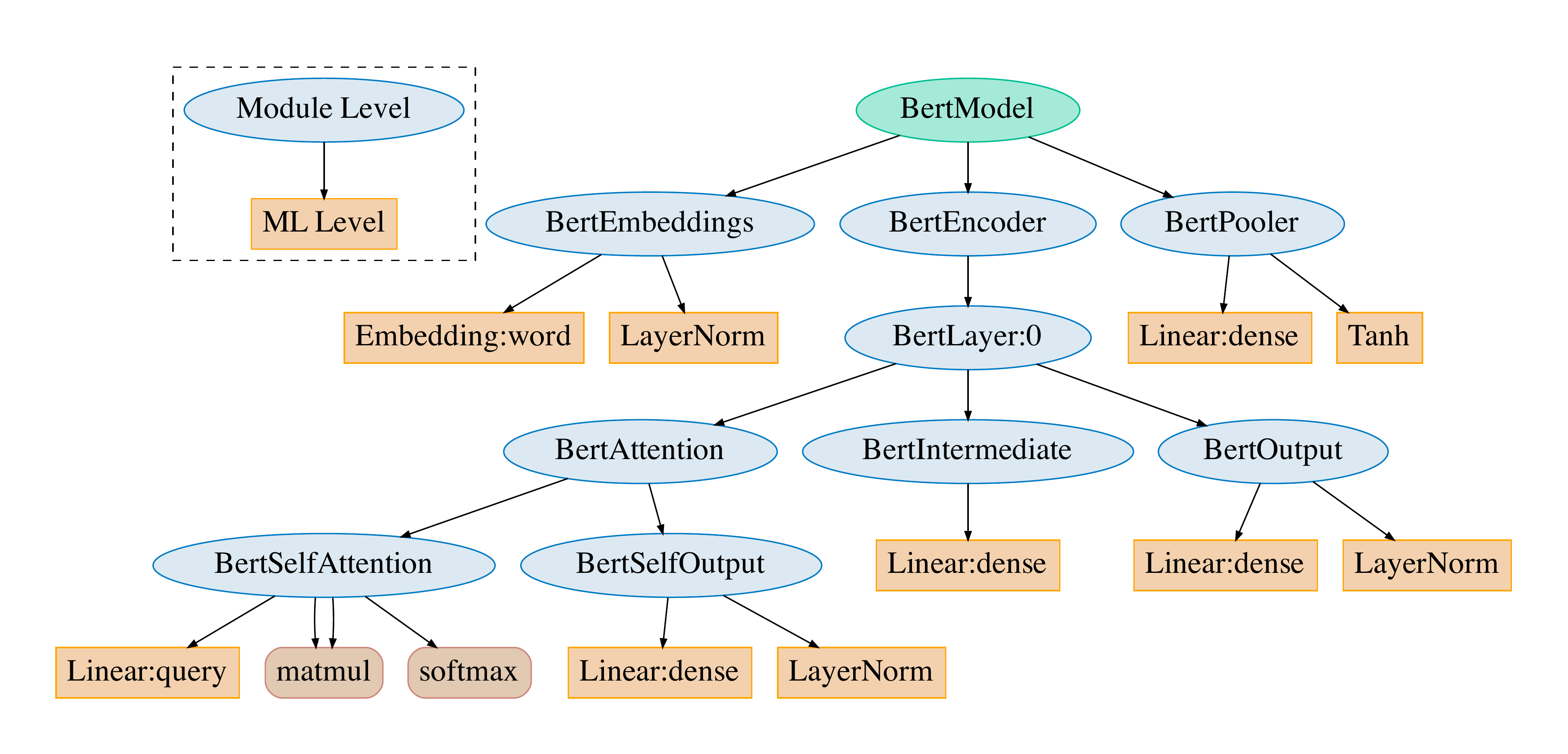}
		\vspace{-1.5em}
		\caption{A tree view of a 1-layer BERT model. The yellow rectangle nodes stand for basic machine learning (ML) level operations. The brown rectangle nodes are also ML level which are non-parametric (i.e., has no trainable parameters). The ML level operations are model-agnostic and provided by machine learning software framework. The light blue oval nodes denote model-specific operations that reflect the architectural semantics given by the model developer, for example BertSelfAttention was designed to transform input sequence representations by `attending" (weighted combination) to each position of the input sequence. }
		\label{fig:model-viz}
		\vspace{-1em}
\end{figure*}

Third, to evaluate \sys, we create an evaluation dataset with ground-truth energy measurements for multiple Transformer-based models at all levels in the model tree abstraction. Evaluations show that \sys\ is more accurate -- with an average model-level energy error of $5\sim7$\% 
compared against the ground-truth, while existing software-based method~\cite{strubell2019EnergyPolicy} has over 55\% error. 
The module-level energy errors are also substantially small showing that \sys\ is both accurate and interpretable. Last, we also conduct multiple analyses that show the utility of \sys\ for interpretable energy predictions.


\section{Related work}



Over the last couple of years, there has been increased interest in the energy consumption of NLP models, starting with the work by Strubell et al.~\cite{strubell2019EnergyPolicy}. This work, and a follow up software framework called \textit{experiment-impact-tracker}~\cite{henderson2020SystematicReporting} tracks the resource (i.e., CPU, GPU, memory) utilization of an NLP model and predicts energy consumption as a function of resources. 
However, our previous study shows that this type of resource utilization only modeling can be highly inaccurate~\cite{cao-etal-2020-towards}. This is in part due to the complex relationship between resource utilization and energy consumption. Further, there are other activities that are not accounted via resource utilization such as data movement in GPU memory which can also cause significant energy footprint~\cite{chen2016EyerissSpatial,boroumand2018GoogleWorkloads}.

Other works~\cite{zhou2020HULKEnergy,schwartz2019GreenAIa} report the energy numbers through alternate metrics including dollar cost or in terms of floating point operations. However, these do not directly map to the energy consumption. Energy prediction of applications on mobile devices is a well-studied topic in the systems community~\cite{pathak2011Finegrainedpower,pathak2012Whereenergya,yoon2012AppScopeapplication,cao2017DeconstructingEnergy} but these work require fine-grained understanding of the application. None of the existing systems  predict energy for NLP applications.

\section{Interpretable Energy Prediction}
In this section we first state our design goals, motivate the abstraction, and problem formulation for interpretable energy prediction.

\subsection{Design Goals} 
We design the energy prediction model with three design goals: (i) \emph{accurate} prediction while incurring low profiling overheads; high overheads when measuring runtime resource utilization can hide the true energy costs of the NLP model, (ii)  \emph{provide interpretable energy analysis} of the components inside the NLP model, especially for analyzing energy bottlenecks; (iii)  \emph{extensible and generalizable}, in the sense that, they are trained once but can work on unseen NLP models to remain useful as new models emerge.

\subsection{Model Tree Abstraction}
To achieve the above goals, we first need a representation of the NLP model that is at a suitable abstraction both from interpretability and generalization standpoints.

On the one hand, using only low-level abstractions such as the math operations 
can help with easy generalization to new models as their units are basic math (or other compute) operations that are building blocks of any model.
However, they lack interpretability since they don't directly convey the model architecture semantics. For example, a BERT~\cite{devlin2019BERTPretraining} model has matrix multiplications in both the self-attention and feed forward layers. Only having the energy of each matrix multiplication alone, without knowing which higher level logic units (i.e., either self-attention or feed forward layer) they belong to, does not help analyze if they are the bottlenecks for that particular unit.
On the other hand, high-level abstractions preserve the architecture semantics and are interpretable for practitioners, but they don't easily generalize to unseen models that may not have the same modules used for training.

Instead, we use a model tree abstraction that represents the model nodes in three-levels: math level, machine learning (ML) level and module level. Math level nodes are a finite set of mathematical operations (like addition, subtraction, matrix multiplication etc); they form model-agnostic ML level nodes (such as Linear, LayerNorm etc.), which further can be used to construct complex module level nodes. Module level nodes are groups of lower ML level node operations that reflect the logic units of the NLP algorithms defined by model authors. The model tree abstraction is such that each parent node captures computation of all of its children nodes. Figure~\ref{fig:model-viz} shows an example of a one-layer BERT~\cite{devlin2019BERTPretraining} model (omitted math level nodes). 
The execution of the model tree nodes can be in parallel, but current systems have a fixed sequential order for executing the sibling nodes. 
In this work, we only focus on sequential execution.
Note that the model tree doesn't capture the order of execution. E.g., \texttt{BertOutput} appears right after \texttt{BertIntermediate} in BERT's computation graph, but here they'll be represented as siblings of the same parent \texttt{BertLayer:0}, and their energy will be treated separately. The parent node \texttt{BertLayer:0} encapsulates the energy and computation of its children node \texttt{BertIntermediate}, \texttt{BertOutput}, and \texttt{BertAttention}, in no particular order.

\subsection{Problem Definition}
With this new model tree abstraction, we formally state the problem of interpretable energy estimation of a NLP model. 
Given a model tree abstraction of a NLP model $\mathcal{M}$ consisting of a set of nodes $\mathcal{N}=\{n|n_{ml} \cup n_{mod}\}$ ($n_{ml}$ is the set of ML level nodes, $n_{mod}$ is the set of module level nodes), for an input size $\mathcal{I}$ (a pair of batch size $b$ and sequence length $s$) \footnote{The batch size and input sequence length together decide the amount of input data to the model, therefore, they both affect the model energy consumption.}, we can predict the energy $E_n$ for every node $n$ in the model tree. 
The energy of root node is the energy for the entire model. 

\begin{figure*}[ht!]
  \centering
		\includegraphics[width=0.9\linewidth]{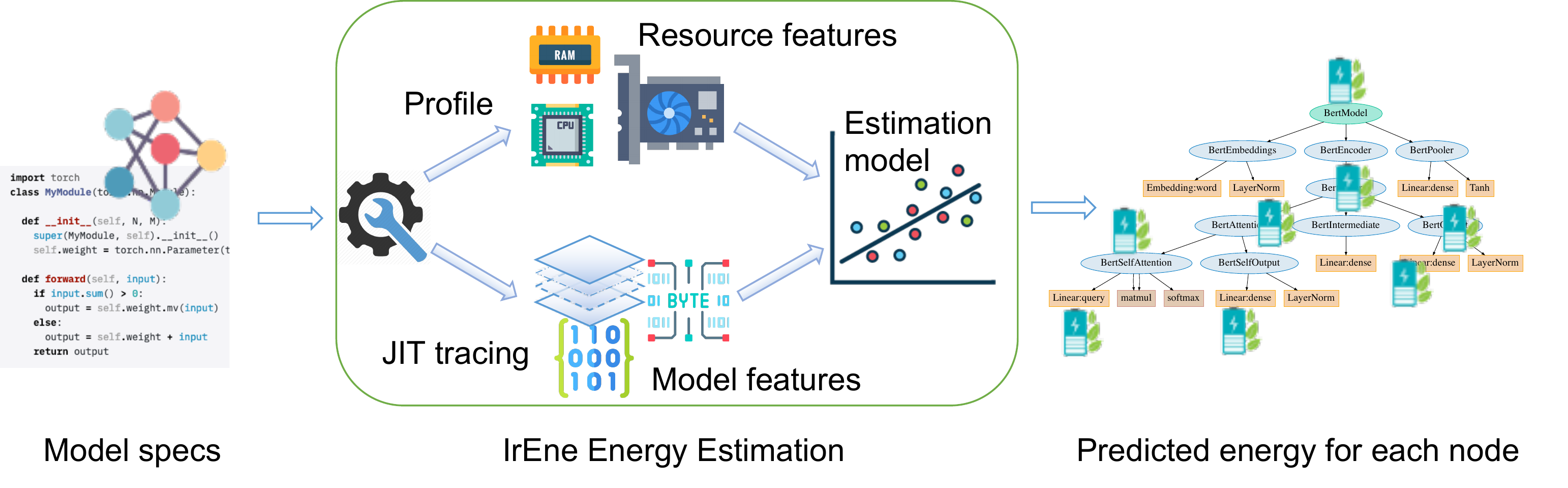}
		\vspace{-0.15in}
		\caption{\sys\ works by taking model specifications (for example, model code) as inputs and extracting a model tree representation using code instrumentation and run-time tracing. \sys\ then runs the model once on a given hardware and feeds resource profiles combined with the model computation features into a regressor to predict the energy of the entire model tree representation. The root of the tree represents the energy of the entire NLP model and each child node represents the energy of different modules/ML operators that make up the model.}
		
		\label{fig:arch}
	
\end{figure*}
\section{Interpretable Prediction with \sys\ }
Figure~\ref{fig:arch} shows the \sys\ architecture. \sys\ takes the user-specified model and builds an energy predictor for a target hardware device.
The model is run once on the target hardware and the runtime resource utilization is logged. During this run, \sys\ uses code instrumentation and just-in-time (JIT) run-time tracing to break down the model into sub-components, and extracts a model tree representation (see details in \S\ref{sec:impl}). 

\sys\ then provides \textit{interpretable energy analysis} by predicting the energy for every node in the model tree in a bottom-up fashion. At the leaves, where the nodes correspond to the ML primitives, \sys\ uses separate regression models for each type of ML primitive (e.g., one regressor for Linear Layer, another for LayerNorm etc.). For the intermediate nodes, their energy is predicted recursively using a single regressor that makes a weighted combination of the predicted energy values from its children. For both types of regressors,  they use features that are derived from resource utilization (e.g. cpu utilization) and generalized node features (e.g. size of inputs) enabling accurate multi-level energy prediction.

\sys\ represents higher-level modules via \textit{generalizable features} and the ML primitives. Even if the intermediate modules are model-specific (e.g. BertSelfAttention), the features are general, allowing \sys\ to predict energy of unseen models.

The \sys\ model is trained using ground-truth energy measurements of ML primitives and a handful of NLP models; we use a highly accurate hardware power monitor to measure ground truth energy (\S\ref{sec:impl}). Of course, one can use the power monitor to measure energy directly at runtime. However, this is cumbersome and requires physical access to the device which is not always feasible with cloud-based deployments. 
Further, the hardware meter only measures the total energy, which is not interpretable in terms of its components. 

\subsection{Multilevel energy prediction}


At the leaf-level, the energy prediction problem requires predicting the energy of ML primitives. As an offline step, \sys\ first enumerates all relevant ML primitives and builds a specialized regressor for each primitive by training over ground truth data. In some cases, model developers can define their own ML primitives. We extract information about such custom primitives from the JIT trace.  

Formally, for a leaf node $n$ with ML primitive $i$, we predict the energy of the node as:

\begin{align}
    P_e^{ML_i}(n) = \mathbf{W}_i*\text{feat}(n)+b_i\label{eqn:ml-regression}
\end{align}

using primitive specific parameters $\mathbf{W}_i$ the weight vector and $b_i$ the bias. We learn these parameters using a mean squared error loss between predicted $P_e(n)$ and ground-truth energy $G_e(n)$.

Our hierarchical tree representation gives a naturally interpretable way of propagating this prediction through the tree. Since each node represents total computation of its children nodes, the total energy from children nodes should also roughly correspond to that of the parent node. Formally,

\begin{align}
%
    P_e(n) &= \sum_{c \, \in \, child(n)} P_e(c) \enspace \text{if n is non-leaf} \nonumber\\
    &= P_e^{ML_i}(n) \quad \quad \quad \text{if n is leaf}
              \label{eqn:predicted-sum}
\end{align}

We call this baseline prediction model \textbf{PredictedSum}. This model is interpretable but naively summing up the energy values accumulates error going up the tree and results in noisy module-level predictions. To account for this, we use a \textit{weighted} sum of child node energy, where the weights are learnt using node features. Formally,

\begin{align}
    P_e(n) &= \hspace{-1em}\sum_{c \, \in \, child(n)}\hspace{-1em} \alpha(c)* P_e(c) \enspace \text{if n is non-leaf}\nonumber \\
                   &= P_e^{ML_i}(n) \quad \quad \quad \quad \text{if n is leaf} \nonumber \\
    \alpha(c)      &= 1 + \text{tanh}(\mathbf{W}*\text{feat}(c) + b)/\tau \label{eqn:end2end-predicted-sum}
\end{align}
where $\mathbf{W}$ and $b$ are parameters and $\tau$ is a hyperparameter. Unlike ML primitives, here we have a single regressor with one set of weight vector ($\mathbf{W}$) and bias scalar ($b$) parameters across all non-leaf nodes of any type. Note that this single regressor doesn't predict node's energy directly, but determines how much the predicted energy from its child node should be scaled before summing the children node energy. It does this recursively starting from the root, and hence encodes tree structure in its computation. We do not learn node-specific regressors because that does not allow generalizing to new models that may have different modules than the ones during training.


Since the method is essentially calibrating the sum of the energy values, regularizing the model so that the computed weights on the energy values to be around 1 helps the learning. We do this by equation \ref{eqn:end2end-predicted-sum}, which makes the range of computed weights, $\alpha(c)$ to be within $1 \pm \tau$.
To supervise this model, we use the ground-truth energy from all the non-leaf nodes, and we train it in an end-to-end fashion. Formally,

\begin{align}
    loss(n) &= \sum_{s \, \in \, subtree(n)} \frac{\big(P_e(s) - G_e(s)\big)^2}{G_e(s)^2} \label{eqn:loss}
\end{align}

We scale the mean squared error with ground-truth energy, since scales of energy at different levels of the tree are vastly different. We refer to this model as the \textbf{End2End} regressor, since the error signal in energy prediction of any node backpropagates through the whole subtree. We use this training scheme in \sys. In our evaluation (section \ref{sec:eval}), we perform an ablation study to show why the tree structure and the end-to-end regressor is crucial for accuracy.


\subsection{Featurization}
\label{sec:featurization}
We design two categories of energy-relevant features in \sys\ : (i) the model features that reflect hardware-independent compute and memory information, and (ii) the resource features that capture how the models use hardware resources and cause energy activities. Table~\ref{tab:features} shows the features used in \sys. For the model description related information, we use features that characterize the compute, memory, and size of input etc. These are features that are independent of the underlying hardware. For resource features, we use utilization, usage and clock speed of hardware components including CPU, memory and GPU.  Note that these two sets of features are extensible, meaning that one can add more either hardware-specific features or new model features. 
See Appendix sections \ref{sec:model-features} and  \ref{sec:resource-features} for details on how we obtain these features. 
\begin{table}[t]
\centering
\begin{tabular}{l}
\toprule
batch\_size : batch size                                 \\
seq\_len    : \# of input tokens                         \\
flops       : floating point operations (unit: million)  \\
mem\_bytes  : memory read and write (unit: MiB) \\ \midrule
cpu\_util         : CPU utilization (unit: \%)                 \\
mem\_usg         : memory usage (unit: \%)                    \\
gpu\_util         : GPU processor utilization (unit: \%)       \\
gm\_usg    : GPU memory usage (unit: \%)                \\
g\_clk   : GPU processor clock speed (unit: MHz)   \\
gm\_clk : GPU memory clock speed (unit: MHz)  \\
latency : inference latency (unit: s)                \\
gpu\_energy : GPU driver energy (unit: joule)   \\ \bottomrule
\end{tabular}
    \caption{\label{tab:features}Features used for energy estimation in \sys.}
\end{table}

\section{\sys\ Evaluation}
\label{sec:eval}
Our evaluation is aimed at measuring the accuracy of \sys\ relative to ground truth and the state-of-the-art. We show the \sys\ only causes 5-7\%  error for the model energy prediction. We also show that for a given Transformer model, \sys\ can be used to  find the energy bottlenecks and analyze the energy versus task performance trade-offs.

\subsection{Setup}

{\noindent \bf Target Hardware:} we use 2 GPU-equipped desktop PCs as the target hardware for running our models. See Table \ref{tab:setup} for details.

\begin{table}[t]
  \centering
  \setlength\tabcolsep{2pt}
  \begin{tabular}{@{}lll@{}}
  \toprule
  Specification & \textbf{PC1}     & \textbf{PC2}   \\ \midrule
  CPU           & Intel i9-7900X   & Intel i7-6800K \\
  Memory        & 32 GiB           & 32 GiB         \\
  GPU           & 2$\times$  GTX 1080 Ti & 2$\times$ GTX 1070   \\
  GPU Memory    & 11.2 GiB per GPU & 8 GiB per GPU  \\
  Storage       & 1 TiB SSD        & 1 TiB SSD      \\ \bottomrule
  \end{tabular}
  \caption{Target hardware specifications. }
  \vspace{-1em}
  \label{tab:setup}
\end{table}

{\noindent\bf Software and models:} We perform inference in Transformer models using PyTorch~\cite{paszke2019PyTorchImperative} v1.7 through the HuggingFace Transformers~\cite{wolf2020HuggingFaceTransformers} library. The six models we study are --- BERT-base~\cite{devlin2019BERTPretraining}, RoBERTa-base~\cite{liu2019RoBERTaRobustly}, DistillBERT~\cite{sanh2020DistilBERTdistilled}, DistilGPT2~\cite{sanh2020DistilBERTdistilled,radford2019language}, OpenAI GPT~\cite{radford2018improving} and GPT2~\cite{radford2019language}.

{\noindent \bf Software-based Measurement Baseline:} For comparisons, we use the software-based energy measurements provided by the {\em experiment-impact-tracker}~\cite{henderson2020SystematicReporting} which estimates energy as a function of the GPU, CPU, and memory utilization. The method computes energy by aggregating resource usage as follows: $e_{total} = PUE \sum_p (p_{dram}e_{dram} + p_{cpu}e_{cpu} + p_{gpu}e_{gpu})$, where $p_{resource}$ \footnote{$resources$ can be $dram$, $cpu$, $gpu$} are the percentages of each system resource used by the attributable processes relative to the total in-use resources and $e_{resource}$ is the energy usage of that resource. The constant for power usage effectiveness (PUE) compensates for extra energy used to cool or heat data centers. 

\subsection{Dataset and Evaluation Methodology}

For each model, we obtain the model tree and for each node in it, we associate ground-truth energy measurements using the power monitor and its resource features using low-overhead logging (Section ~\ref{sec:impl}). For each node we run it repetitively for 20 seconds, since it often takes a very short time for one run (e.g. from 0.1 to 100 millisecond). 
We repeat this process for five rounds (the variations are within <1\%) and record the average energy as the ground-truth for the node. We use 1 GPU to run all experiments. 
We record the start and end timestamp of the model program, and extract the energy values by comparing and aligning the timestamps from the resource profiler logs and power monitor logs. 


{\noindent\bf Ground Truth Energy:} 
We measure ground truth energy using a  emonPi power monitor~\cite{hudsonEmonPi} which is open source. The emonPi uses a clip-on CT sensor to monitor the energy consumed by the computer which records the passthrough current and voltage every 170 ms. This allows us to accurately measure the power draw at a sub second granularity. 
We obtain current, voltage, and timestamp values from the power meter's built-in serial port. The energy ($e$) consumed during a time period is then calculated using the sampled current $(I_t)$ and voltage $(V_t)$ values in that period: $e = \sum_{t} V_t I_t$.

To guarantee the consistency and reliability of the hardware energy measurements, we cool down the PCs after each experiment finishes to avoid potential overheating issue that can cause subsequent energy distortions. We  measure the standby power consumption (when the CPU load is $<0.1$\%) and ensure before running the experiments that the PC does not draw more than the standby power. Further, no other application is running during our experiments. 

To understand the scale of energy usage, Table~\ref{tab:example-energy} shows the estimated energy consumption (in kWh) using our ground truth measurement. We also show the cost of answering one million queries (in USD) when using a BERT-base model in a reading comprehension (over one passage), and in an end-to-end setting (over 150 passages) ignoring retrieval compute. For reference, Google search handles millions of queries every minute~\cite{HowManyGoogle2019}.

\begin{table}[th!]
    \centering
    \small
    \setlength\tabcolsep{6pt}
    \begin{tabular}{lll}
        \toprule
        Use Case &  \makecell{Energy/1M  \\ Qns (kWh)} & \makecell{Cost/1M  \\ Qns (USD)} \\ \midrule
        \hspace{-0.25em}QA over a single passage & 161 & 21.24  \\
        \hspace{-0.9em}\makecell{QA over 150 passages \\ (ignore search/retrieval)}     & 24,000 & 3,165\\
        \bottomrule
        \end{tabular}
    \caption{Example energy for BERT-base QA models using batch size 16 and sequence length 256 on PC1 using one GPU. The cost is estimated at $13.19$ cents per kWh. \footnotemark}
    
    \label{tab:example-energy}
\end{table}

\footnotetext{based on the US national average as of May 2021 according to  \url{https://www.electricchoice.com/electricity-prices-by-state}.}

{\noindent\bf Energy Dataset:} To evaluate the energy prediction, we create a dataset that cover a wide range of input sizes for the six studied Transformer models and the 24 BERT model variants~\cite{turc2019WellReadStudents}.
Each instance in the dataset can be of type ML, Module or Model level and is associated with features shown in Table \ref{tab:features} and hardware measured energy. We show the statistics of the dataset for BERT-base, DistilBERT and GPT2 in Table \ref{tab:bert-data-statistics}.

\begin{table}[t!]
    \centering
    \setlength\tabcolsep{3pt}
\begin{tabular}{@{}llll@{}}
\toprule
Quantity        & BERT-base & DistilBERT & GPT2 \\ \midrule
\# ML Nodes     & 3864      & 1932       & 2997 \\
\# Module Nodes & 2100      & 560        & 972  \\
\# Model Nodes  & 28        & 28         & 28   \\
\# Tree Depth   & 6         & 5          & 4    \\ \bottomrule
\end{tabular}
    \caption{Energy dataset statistics for BERT-base, DistilBERT and GPT2  model. For each model, we construct 28 trees (model nodes) with batch sizes from 8 to 32 with a step of 8, and input sequence lengths from 32 to 256 with a step of 32. We associate features and ground-truth energy for each node in these trees. 
    }
    \label{tab:bert-data-statistics}
\end{table}

{\noindent\bf Energy Error Metric:} We measure the energy error percentage as $100 \times |PE-GE|/GE$, where $PE$ is the predicted energy and $GE$ is the ground truth energy.

\subsection{Energy Prediction Results}

We compare \sys  with the existing software measurement methods~\cite{strubell2019EnergyPolicy,henderson2020SystematicReporting}. We apply their method directly for all the models in our dataset. Note that their method is a fully-defined estimation model with a fixed set of parameters without any training. 
For \sys\ experiments, we report cross-validated evaluation on the energy prediction dataset --- leaving data from one model out of training set and evaluating on it, and then repeating the same for all the models.

{\noindent\bf \sys is accurate}
Table~\ref{tab:energy-error} shows the energy prediction errors at the model-level for all the models on the two PCs. The existing software-based baseline method from \citet{strubell2019EnergyPolicy} incurs large energy prediction errors of over 50\%. 

\begin{table*}[ht!]
    \centering
    \small
    \setlength\tabcolsep{3pt}
\begin{tabular}{@{}lllllllll@{}}
\toprule
Machine & System                                              & BERT-base & DistilBERT & RoBERTa-base & GPT2 & DistilGPT2 & OpenaiGPT & Average \\ \midrule
PC1     & \citealp{strubell2019EnergyPolicy} & 57.9      & 56.3       & 62.5         & 62.6 & 55.9       & 61.8      & 57.8    \\
        & \sys                 & \textbf{5.8}       & \textbf{11.6}       & \textbf{7.1}          & \textbf{3.5}  & \textbf{2.2}        & \textbf{2.7}       & \textbf{5.5}     \\ \midrule
PC2     & \citealp{strubell2019EnergyPolicy} & 55.1      & 52.6       & 58.9         & 54.6 & 49.8       & 60.6      & 55.6    \\
        & \sys                 & \textbf{10.0}      & \textbf{9.4}        & \textbf{7.1}          & \textbf{6.1}  & \textbf{4.9}        & \textbf{5.9}       & \textbf{7.2}     \\ \bottomrule
\end{tabular}
    \caption{Energy Prediction Errors at Model level: Comparing \sys\ and a software measurement baseline for the two PCs. \sys is significantly more accurate than \citealp{strubell2019EnergyPolicy}.
    } 
    \label{tab:energy-error}
\end{table*}

\begin{figure}[t!]
	\begin{center}
		\includegraphics[width=0.8\linewidth]{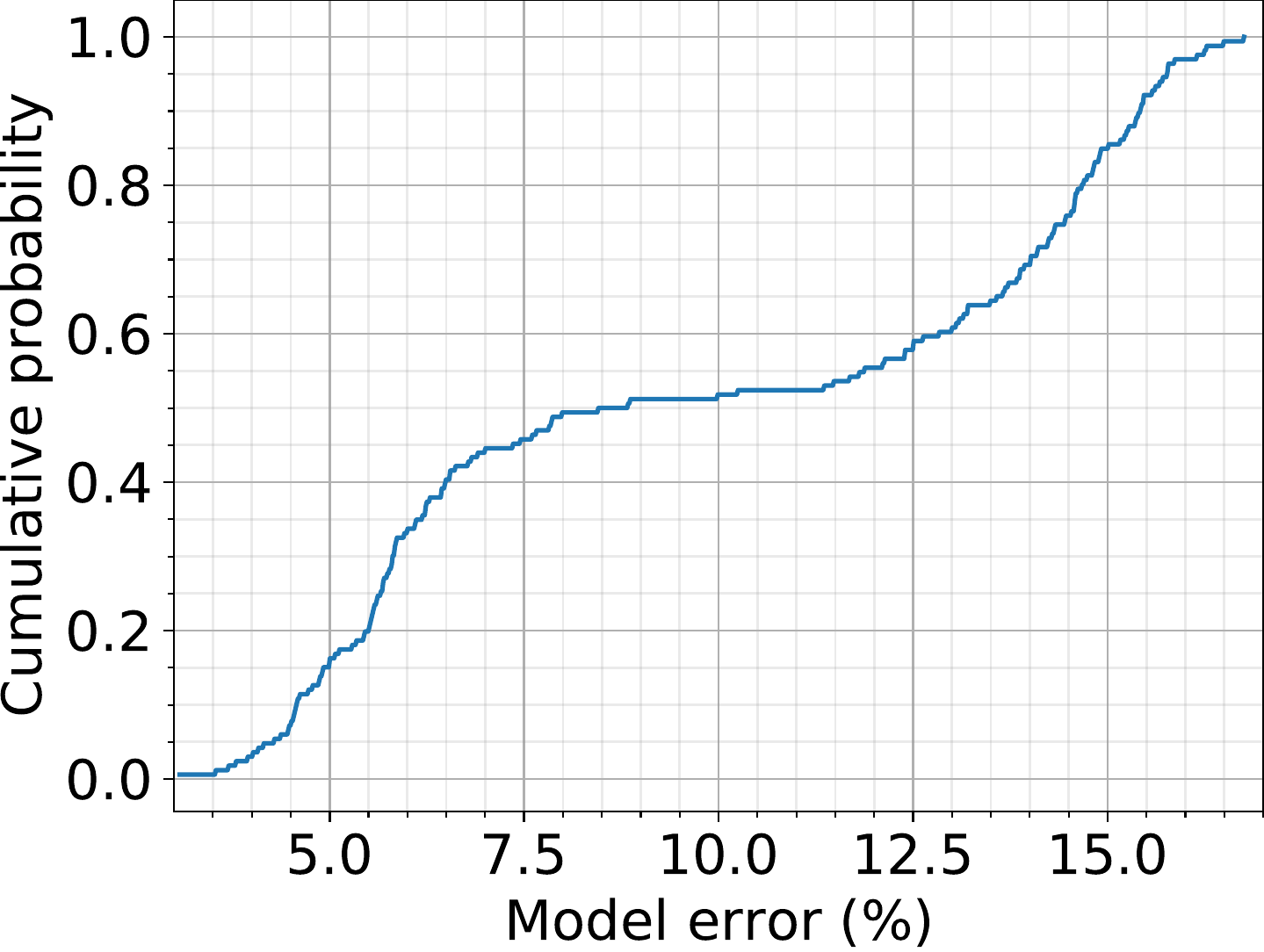}
		\caption{\label{fig:error-cdf}{The CDF of model's predicted energy errors. We see that for 99\% of the cases, the error is under 16\%}}
		\vspace{-1.5em}
	\end{center}
\end{figure}

\sys\ on the other hand incurs substantially lower errors,  with at most 7.6\% errors across the models, showing its value for reliable and accurate energy analysis. As seen from the cumulative distribution function for the model errors in Figure~\ref{fig:error-cdf}, all of \sys's errors are below 17\% and nearly half of its errors are below 10\%. We note here that our leave-one-model-out cross validation specifically evaluates the generalizability of \sys. 

\noindent{\bf ML and Module Levels Errors are also low.}
Table~\ref{tab:ml-level-energy-error}, ~\ref{tab:module-level-energy-error} show a break down of the \sys\ errors at the ML and module levels respectively. Accurately predicting ML level energy is key to accurate predictions for at the module level and higher, as the errors will accumulate up the model tree in \sys. It turns out that we can indeed predict ML level energy with high-levels of accuracy --- errors are lower than 1\%, providing reliable values for the module level predictions. Note that even unseen models (ie ones evaluated in the test partition) will be made up of the same set of ML primitives (perhaps with different input and batch sizes). The results here cannot be directly generalized to unseen ML-primitives. 
Module level errors are higher and vary in range (5.4\% to 16.7\%) across different models. 
Module level errors also turn out to be higher than the model level errors. This is mainly because the module level errors are averages across all intermediate module level nodes in the model tree; some modules might have bigger errors, but these get calibrated by our \textbf{End2End} energy regressor. We further characterize these effects in \sys\ ablation and validation analysis.

\begin{table*}[ht!]
    \centering
    \small
    \setlength\tabcolsep{7pt}
    \begin{tabular}{@{}llllllll@{}}
        \toprule
    Machine &      BERT-base & DistilBERT & RoBERTa-base & GPT2 & DistilGPT2 & OpenaiGPT  & Average\\ \toprule
   PC1   & 5.37      & 5.93       & 5.44   & 14.92 & 14.73 & 13.98 & 8.54 \\
   PC2   & 6.78      & 7.96       & 6.69   & 16.65 & 16.41 & 16.07 & 10.16 \\ \bottomrule
        \end{tabular}
    \caption{Energy Prediction Errors at module levels using \sys\ on two PCs. Note that in  Table~\ref{tab:soft-module-level-energy-error} at the appendix, we also show a subset of the module level energy errors using \citealp{strubell2019EnergyPolicy}.}
    \label{tab:module-level-energy-error}
\end{table*}

\begin{table*}[ht!]
    \centering
    \small
    \setlength\tabcolsep{8pt}
    \begin{tabular}{@{}lllllllll@{}}
        \toprule
    Machine & Embedding & LayerNorm & Linear & Tanh & MatMul & Softmax  & Conv1D & Average\\ \toprule
    PC1 & 0.65      & 0.89       & 0.60   & 0.82  & 0.61  & 1.0   & 0.58 & 0.70 \\
    PC2 & 0.38      & 0.66       & 0.55   & 0.43  & 0.43  & 0.67  & 0.41 & 0.53 \\ \bottomrule
        \end{tabular}
    \caption{Energy Prediction Errors at ML levels using \sys\ on two PCs. 
    Note that the evaluation for these operation-specific (eg. Embedding) regressors is done using the leave-one-model out setting as before.}
    \label{tab:ml-level-energy-error}
\end{table*}



\subsection{\bf Feature Ablations} Table~\ref{table:feature-ablation} shows the contribution of model and resource features in \sys\ energy prediction. We observe that resource features provide most of the benefits for energy estimation \sys\ for all levels, confirming that resource information is important for energy prediction. Model features do not reduce ML level error because the error is already small, but they help further reduce the prediction errors for module and model levels and combining model and resource features together brings the average estimation errors further down to 8.5\% and 5.5\%. 

\subsection{Modeling Ablations}
To understand the impact of learning and the  architectural choices of aggregating ML level energy into module level energy in \sys\ affect the model accuracy, we build three (ablated) models:

\noindent \textbf{Is end-to-end learning necessary?} To test this, we build a \textbf{StepWise} regressor that simply learns to predict the energy of parent node from the ground-truth energy of its child nodes at the training time. At the test time, it uses predicted energy generating predictions from ground up. 

\begin{align}
    P_e(n) &= \sum_{c \, \in \, child(n)} \alpha(c)* G_e(c) & \text{Training} \nonumber \\
    P_e(n) &= \sum_{c \, \in \, child(n)} \alpha(c)* P_e(c) & \text{Testing}
\end{align}

Here, $\alpha(c)$ and loss are still as defined in equation \ref{eqn:end2end-predicted-sum} and \ref{eqn:loss} respectively. However, unlike the \sys (\textbf{End2End}) regressor, the errors in the prediction of root node, do not backpropogate to its prediction of descendant nodes i.e. there is no end-to-end training.

\noindent \textbf{Is tree-structure necessary?} To test this, we build an \textbf{Unstructured} regressor that ignores the tree structure completely, and directly predicts the energy from the feature representation of nodes (Module and Model level) using linear regression as in equation (\ref{eqn:ml-regression}). Unlike ML-level regressor though, here we need to use single set of parameters for common across the nodes.

\begin{table}[t!]
  \centering
  \setlength\tabcolsep{2.5pt}
  \small
  \begin{tabular}{lcccccc}\toprule
                            & ML & Module & Model \\
      \midrule
      \sys\                         & 0.70   &  \textbf{8.54} & \textbf{5.52} \\
      \midrule
      \quad w/o resource features   & 5.76  & 11.54 & 7.08 \\
      \quad w/o model features      & \textbf{0.63}  &  8.87 & 7.32 \\
      \bottomrule
  \end{tabular}
  \caption{Energy Prediction Errors of \sys with ablated features. Both model and resource features help the \sys's performance at model and module levels, while resource features are sufficient for ML-level.}
  \label{table:feature-ablation}
\end{table}

\begin{table}[t!]
  \centering
  \setlength\tabcolsep{12pt}
  \small
  \begin{tabular}{lccccc}\toprule
                              & Module & Model \\
      \midrule
      \sys\ (End2End)         & \textbf{8.54}   & \textbf{5.52} \\
      \midrule
      \qquad StepWise         & 9.28   & 14.84 \\
      \qquad PredictSum       & 16.4   & 17.69 \\
      \qquad Unstructured     & 278.0  & 39.79 \\
      \bottomrule
  \end{tabular}
  \caption{Energy Prediction Errors of \sys using different module/model level regressors on PC1. Tree structure of the regressor crucial, and end-to-end optimisation on tree helps \sys to get lower errors.}
  \vspace{-1.5em}
  \label{table:architecture-ablation}
\end{table}

\noindent \textbf{Is learning necessary?} To test this, we use the \textbf{PredictedSum} model (equation \ref{eqn:predicted-sum}). Recall this model also aggregates energy predictions over the tree-structure but has no parameters to train.

Table~\ref{table:architecture-ablation} shows the ablation of \sys\ with respect to different algorithmic choices of the module level energy aggregation. First, we find that the regressor that ignores the tree structure (\textbf{Unstructured}) performs significantly worse than all other regressors that do consider it. 
Interestingly, learning without structure even performs worse than \textbf{PredictedSum} regressor that naively adds child energy without any learning, highlighting the importance of tree-structure. Further, learnt weighted sum outperforms \textbf{PredictedSum} regressor. 
In particular, \textbf{End2End} regressor performs better than \textbf{StepWise} regressor showing the importance of optimizing on whole tree in an end-to-end fashion.






\subsection{Interpretable Energy Analysis}
In this section, we use the interpretable energy analysis from \sys\ to show energy bottlenecks for given Transformer models, how energy varies for different model architectures, and how it can be used to effectively pick accuracy-energy trade-offs.

{\noindent\bf Finding energy bottlenecks:} We use \sys\ to analyze the energy bottlenecks in Transformer models. For simplicity of analysis, we predict the energy for modules that are immediate parents of the ML level nodes and use it calculate the percentage of energy it contributes to the model overall. 
Table~\ref{tab:pred-bottleneck} shows the energy breakdown of two models: RoBERTa-base and GPT2. We observe that self-attention layers in RoBERTa-base model consume 31\% of the total energy while it is the feed forward layers in GPT2 that consume more than 59\% of the energy. The module level energy breakdown of all models in Table~\ref{tab:pred-bottleneck-apdx} in Appendix~\ref{sec:bottlenecks_apdx}.
We also present the full energy breakdown of the BERT-base model and annotate each node with predicted energy percentage in Figure ~\ref{fig:model-viz-energy} in the Appendix.

\begin{table}[t!]
\small
\setlength\tabcolsep{2pt}
\begin{subtable}[t]{0.45\linewidth}
    \centering
    \begin{tabular}[t]{@{}lc@{}}
    \toprule
    Module        & Energy \% \\ \midrule
    RobertaSelfAttention & \cellcolor{orange!33}31.24            \\
    RobertaIntermediate  & \cellcolor{orange!30}30.57            \\
    RobertaOutput        & \cellcolor{orange!27}28.64            \\
    RobertaSelfOutput    & \cellcolor{orange!9}09.11            \\
    RobertaEmbeddings    & \cellcolor{orange!4}00.41            \\
    RobertaPooler        & \cellcolor{orange!2}00.03            \\ \bottomrule
    \end{tabular}
    \caption{RoBERTa-base}
    \label{tab:roberta-pred-bottleneck}
\end{subtable}
\hfill
\begin{subtable}[t]{0.45\linewidth}
    \centering
    \begin{tabular}[t]{@{}lc@{}}
    \toprule
    Module      & Energy \% \\ \midrule
    MLP         & \cellcolor{red!62}59.13 \\
    Attention   & \cellcolor{red!35}37.94 \\
    LayerNorm   & \cellcolor{red!3}2.84 \\
    Embedding   & \cellcolor{red!2}0.1 \\ 
    & \\
    & \\
    \bottomrule
    \end{tabular}
    \caption{GPT2}
    \label{tab:gpt2-pred-bottleneck}
\end{subtable}
\caption{Module level predicted energy breakdown of two Transformer models. We average the energy of these modules across all input sizes for each model architecture. Self-attention is the energy bottleneck in RoBERTa-base, but for GPT2, the bottleneck is feed forward layers (MLP module). 
}
\label{tab:pred-bottleneck}
\end{table}

{\noindent\bf Task accuracy versus energy tradeoffs:} 

We fine-tune BERT-24 models~\cite{turc2019WellReadStudents} on the Stanford Sentiment Treebank V2 (SST2)~\cite{socher-etal-2013-recursive} using the default examples in the HuggingFace Transformers~\cite{wolf2020HuggingFaceTransformers}  without any hyperparameter tuning. 
We evaluate the accuracy on the dev set of SST2. These models are not part of our energy prediction training data. We additionally exclude BERT-base from training data to show the extensibility of \sys. 



\begin{figure}[t!]
	\begin{center}
		\includegraphics[width=0.85\linewidth]{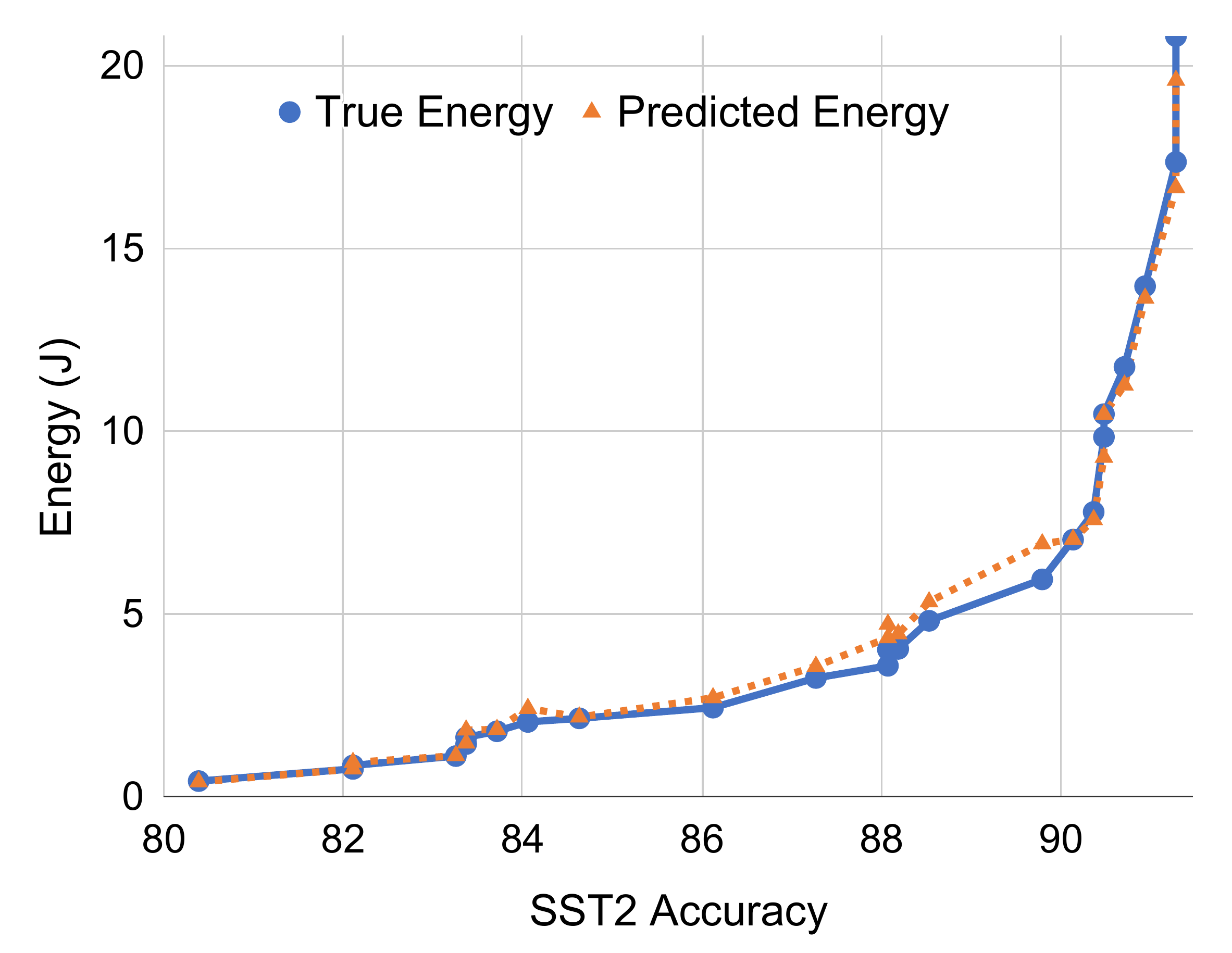}
		\caption{\label{fig:acc-energy}{Ground-truth and predicted energy vs accuracy on SST2 task for BERT-24 models. Energy data is collected with batch size 16 and sequence length 128. Because our energy predictions are accurate, we can use energy consumption vs NLP model accuracy trade-offs to select a model.}}
		\vspace{-2em}
	\end{center}
\end{figure}

Given an energy budget, \sys allows for selection of an optimal architecture that gets the highest accuracy for a task.
In Figure ~\ref{fig:acc-energy}, we see that it is possible for models to use more energy but return lower accuracy than other models which might use less energy. Similarly, given an accuracy target, we can choose an  architecture with the lowest energy use. For example, for a target of 88\% accuracy or above, there are many such models ranging from 4J all the way to 12J. Last, we point out that the trade-off curve based on the predicted energy mirrors that of the ground-truth well enough 
to be used as an accurate proxy.

\section{Discussion}
This work focused on inference energy predictions of Transformers on a target hardware device. The model tree abstraction is general and not tied to Transformer architectures nor to specific deep learning frameworks, it is extensible to other neural networks like LSTM and frameworks like TensorFlow. The abstraction is built from the computational graph and knowledge about the model architecture and underlying software. As long as these are available we can apply our methodology to other architectures as well. 

Predicting the training energy is an important and a more challenging problem. We believe our methodology can be extended. However, it will require tracking the energy of both forward and backward processes and even modeling other aspects training dynamics, for example, time to converge to specific accuracy. 

Scaling to unseen hardware is an important and challenging area that needs further research. It requires both measuring the ground truth energy for a more diverse collection of hardware and designing proper hardware-specific features (i.e., L1 cache size, CPU cores, etc.). We believe IrEne’s methodology can be extended to calibrate software reported energy as a way to scale how we collect ground truths (as weak-supervision). In the future, we plan to study workloads on more hardware to choose proper features that capture the hardware energy differences.

\section{Conclusions}

Energy consumption of NLP models is an important consideration from a cost perspective and increasingly, from an environmental impact perspective as well. Designing energy efficient and cost-effective models requires both accurate and interpretable energy modeling. In this work, we showed that by carefully combining resource utilization with model description based features, we can develop a multi-level energy prediction model that is not only highly accurate but is also able to provide a break-down of how its different components contribute to its overall energy.

\section{Acknowledgement}
\label{sec:ack}
This material is based upon work supported by the National Science Foundation under Grant No 2007362.

\bibliographystyle{acl_natbib}
\bibliography{ref}

\clearpage

\appendix

\section{\sys\ Implementation Details}
\label{sec:impl}

In this section, we provide the implementation details of \sys. \sys\ is implemented for PyTorch \cite{paszke2019PyTorchImperative}, but can be extended to TensorFlow \cite{abadi2016tensorflow} in future.

\subsection{Constructing the model tree}

The first step to extracting the model tree is to run the model on the target hardware. We run the version of the model on HuggingFace Transformers library v4.2.2~\cite{wolf2020HuggingFaceTransformers} for random data of different input sizes. Once run, we have both the execution graph and the JIT trace that provides runtime information. We use existing PyTorch APIs to obtain module level nodes, ML primitives, and the relationships between them, from the execution graph.  In some cases, the NLP model may use customized ML primitives. To extract information about these custom primitives, we combine information from the JIT trace and the execution graph. Once we obtain all the component, we can construct the model tree. 

The following ML primitives are used in Transformers: Linear, LayerNorm, Embedding, BatchNorm1d, Conv1d, MaxPool1d, AvgPool1d, LSTM, Tanh, Conv1D, LogSigmoid, ReLU, Sigmoid, GELU, and LeakyReLU. 
Two custom primitives: matrix multiplications (including torch.matmul, torch.bmm and torch.einsum), softmax (torch.softmax).

\begin{table}[ht]
\centering
\setlength\tabcolsep{50pt}
    \begin{tabular}{@{}lr@{}}
        \toprule
        Machine      & PC1 \\ \toprule
        BERT-base    & 32.54  \\
        DistilBERT   & 62.80  \\
        RoBERTa-base & 13.36  \\
        GPT2         & 24.96  \\
        DistilGPT2   & 35.93  \\
        OpenaiGPT    & 42.37  \\
        Average      & 35.33 \\  \bottomrule
        \end{tabular}
    \caption{Energy Prediction Errors at Module levels using \citealp{strubell2019EnergyPolicy} methodology on PC1.}.
    \label{tab:soft-module-level-energy-error}
\end{table}


\subsection{Model features} 
\label{sec:model-features}
The model features reflect hardware-independent compute and memory information for a given model.
We use the  model execution to extract model features used by \sys\ for energy prediction. We add forward hooks to each node in the model to track the shape and input data of each module and ML primitive. PyTorch hooks only support tuple arguments, but we extend these to also support keyword based arguments. 
The JIT trace contains information about the number of FLOPs and memory bytes for each module and ML primitive. By combining JIT information and the information obtained from our hooks, we get the model features. 

\begin{figure*}[ht]
  \centering
		\vspace{-1em}
		\includegraphics[width=\linewidth]{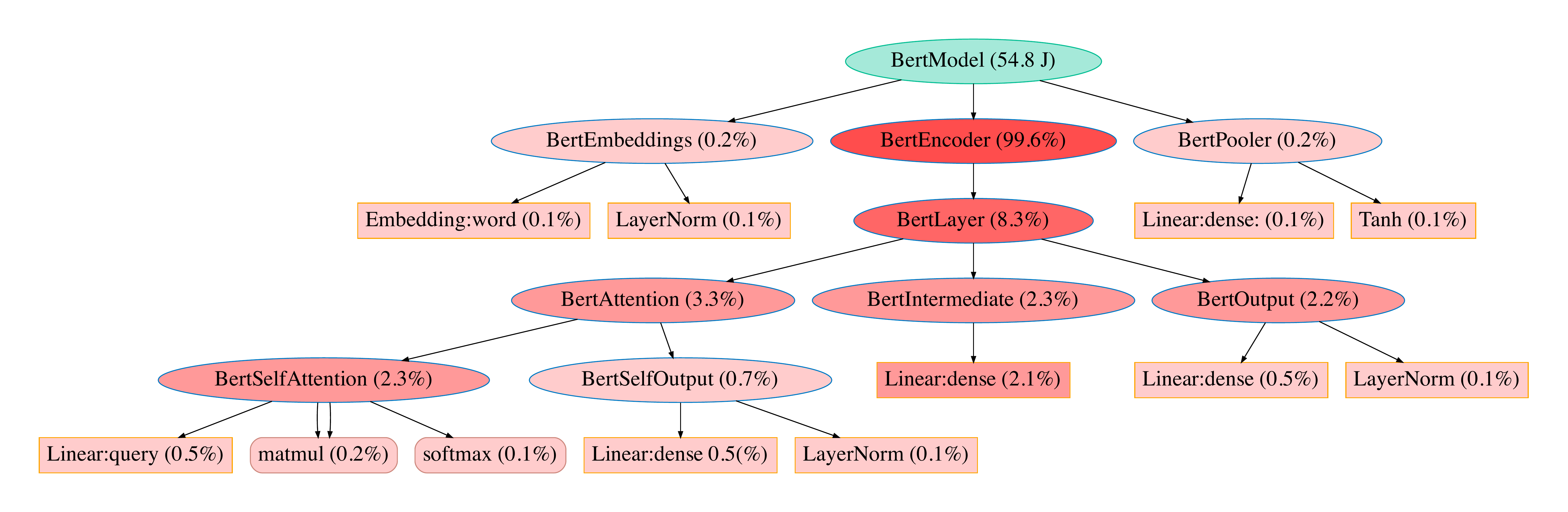}
		\vspace{-1em}
		\caption{Abridged view of a BERT-base-uncased model annotated with predicted energy from our prediction method. The root contains the absolute energy of the model while every other node is annotated with its respective energy percentage share. Darker colors represent nodes that consume a higher percentage of energy. There are 12 BertLayer modules in the actual model. We show just one for brevity. The shown energy is an average of energy of the node across all (batch size, sequence length) models of BERT-base-uncased type.}
		\label{fig:model-viz-energy}
		\vspace{-1em}
\end{figure*}

\subsection{Resource features} 
\label{sec:resource-features}
Resource features capture how the models use hardware resources and cause energy activities.
Existing work~\cite{henderson2020SystematicReporting} uses the OS resource profiler to log the resource utilization of CPU, memory and GPU events. However, this incurs high profiling overhead, and profiling is only done at a low rate of once every second. Instead, to monitor resources, we obtain the CPU utilization by directly reading \verb+/proc/stat+ and memory usage by reading \verb+/proc/meminfo+ via a C program. We simultaneously log the GPU utilization, GPU memory usage, GPU Streaming processor (SM) clock frequency and GPU memory frequency using the Nvidia NVML API~\cite{NvmlUtilization}. To maintain low monitoring overhead, we log resources every 170 ms, resulting in less than 0.5\% increase in CPU utilization and $<15$ MB memory footprint.

Note that both model and resource features are extensible, meaning that one can add more either hardware-specific features or new model features for newer deep learning frameworks or emerging hardware like customized deep learning accelerators. 





\subsection{Regressor Training Procedures}

We've implemented \sys\ using SciKit Learn~\cite{scikit-learn} and PyTorch \cite{paszke2019PyTorchImperative}. We learn linear regressors for ML-level in SciKit Learn \cite{pedregosa2011scikit}, and module and model level regressor in PyTorch, which allows easily optimizing on dynamic tree-structured computation graphs. We use Adam optimizer \cite{kingma2014adam} with 0.001 learning rate. In our experiments $\tau$ in equation \ref{eqn:end2end-predicted-sum} is fixed value of 10. We normalize all the features to have 0 mean and 1 standard deviation, learning mean and standard deviation from the training set and applying it on the test set.



\section{Software Measurements Results}
\label{sec:soft-results}

We use {\em experiment-impact-tracker}~\cite{henderson2020SystematicReporting} to estimate software-based energy measurements for the models at a module level as well as ML level.
Table~\ref{tab:soft-module-level-energy-error} shows the percentage error in software based measurements for module level operations.
We calculate a model's module level error as average percentage error over runs for batch sizes 24 and 38, and sequence length 32 and 128.
Getting granular ML level software energy corresponding to \citet{strubell2019EnergyPolicy} requires modifying the existing framework which is non-trivial.
We leave this to future work.

\begin{table}[ht]
\vspace{-1.35em}
\begin{subtable}[t]{\linewidth}
    \centering
    \begin{tabular}[t]{@{}lc@{}}
    \toprule
    Module        & Energy \% \\ \midrule
    BertOutput        & \cellcolor{orange!33}31.89          \\
    BertSelfAttention & \cellcolor{orange!30}29.26            \\
    BertIntermediate  & \cellcolor{orange!27}27.97            \\
    BertSelfOutput    & \cellcolor{orange!9}09.74            \\
    BertEmbeddings    & \cellcolor{orange!4}00.34            \\
    BertPooler        & \cellcolor{orange!2}00.11            \\ \bottomrule
    \end{tabular}
    \caption{BERT-base}
    \label{tab:bert-pred-bottleneck}
\end{subtable}
\begin{subtable}[t]{\linewidth}
    \centering
    \begin{tabular}[t]{@{}lc@{}}
    \toprule
    Module      & Energy \% \\ \midrule
    MLP         & \cellcolor{red!62}61.41 \\
    Attention   & \cellcolor{red!35}35.70 \\
    LayerNorm   & \cellcolor{red!3}2.79 \\
    Embedding   & \cellcolor{red!2}0.11 \\ 
    \bottomrule
    \end{tabular}
    \caption{OpenAI-GPT}
    \label{tab:gpt-pred-bottleneck}
\end{subtable}
\begin{subtable}[t]{\linewidth}
    \centering
    \begin{tabular}[t]{@{}lc@{}}
    \toprule
    Module Name       & Energy \% \\ \midrule
    FFN                    & \cellcolor{purple!55}57.23                  \\
    MultiHeadSelfAttention & \cellcolor{purple!42}39.46                  \\
    LayerNorm              & \cellcolor{purple!4}2.69                  \\
    Embeddings             & \cellcolor{purple!2}0.62                  \\
    \bottomrule
    \end{tabular}
        \caption{DistilBERT}
        \label{tab:distilbert-pred-bottleneck}
\end{subtable}
\begin{subtable}[t]{\linewidth}
    \centering
    \begin{tabular}[t]{@{}lc@{}}
    \toprule
    Module Name       & Energy \% \\ \midrule
    FFN                    & \cellcolor{blue!55}57.50                  \\
    MultiHeadSelfAttention & \cellcolor{blue!42}39.43                  \\
    LayerNorm              & \cellcolor{blue!4}2.86                  \\
    Embeddings             & \cellcolor{blue!2}0.21                  \\
    \bottomrule
    \end{tabular}
        \caption{DistilGPT2}
        \label{tab:distilgpt2-pred-bottleneck}
\end{subtable}
\vspace{-0.5em}
\caption{Module level predicted energy breakdown of four Transformer models. We average the energy of these modules across all available input sizes for each model architecture. Interestingly, we find that even models with similar architecture have different types of energy bottlenecks. For example, BERT-base has similar architecture to DistilBERT but has different energy bottlenecks.}
\label{tab:pred-bottleneck-apdx}
\vspace{-4em}
\end{table}


\section{Energy Breakdowns}
\label{sec:bottlenecks_apdx}

We show module level predicted energy breakdown of four Transformer models in Table \ref{tab:pred-bottleneck-apdx}, and show an abridged view of BERT-base-uncased tree annotated with predicted energy and distribution in Figure \ref{fig:model-viz-energy}.



\end{document}